\documentclass[letterpaper, 10 pt, conference]{ieeeconf}  

\IEEEoverridecommandlockouts                              

\usepackage{cite}
\usepackage{amsmath,amssymb,amsfonts}
\usepackage{algorithmic}
\usepackage{graphicx}
\usepackage{textcomp}
\usepackage{xcolor}

\title{\LARGE \bf
AI-Enhanced Kinematic Modeling of Flexible Manipulators Using Multi-IMU Sensor Fusion
}

\author{Amir Hossein Barjini$^*$, Jouni Mattila 
\thanks{
   Amir Hossein Barjini is the corresponding author and is with Department of Automation Technology and Mechanical Engineering at Tampere University, Finland. {\tt\small (e-mails: amirhossein.barjini@tuni.fi, jouni.mattila@tuni.fi)}.
}
\thanks{
This work was supported by the Research Council of Finland under the Project "Nonlinear PDE-model-based control of flexible manipulators" under Grant 355664.
}
}

\begin{document}

\maketitle
\thispagestyle{empty}
\pagestyle{empty}

\begin{abstract}
This paper presents a novel framework for estimating the position and orientation of flexible manipulators undergoing vertical motion using multiple inertial measurement units (IMUs), optimized and calibrated with ground truth data. The flexible links are modeled as a series of rigid segments, with joint angles estimated from accelerometer and gyroscope measurements acquired by cost-effective IMUs. A complementary filter is employed to fuse the measurements, with its parameters optimized through particle swarm optimization (PSO) to mitigate noise and delay. To further improve estimation accuracy, residual errors in position and orientation are compensated using radial basis function neural networks (RBFNN). Experimental results validate the effectiveness of the proposed intelligent multi-IMU kinematic estimation method, achieving root mean square errors (RMSE) of 0.00021~m, 0.00041~m, and 0.00024~rad for $y$, $z$, and $\theta$, respectively.
\end{abstract}

\section{Introduction}

Flexible manipulators, characterized by lightweight construction and low energy consumption, have attracted considerable attention in diverse domains such as aerospace~\cite{feliu2022lightweight}, industrial automation, and medical robotics~\cite{zhang2020review}. However, their inherent elasticity and the resulting vibrations introduce strong nonlinearities into the kinematic behavior \cite{wang2025adaptive}, which is further influenced by factors such as payload, motion speed, and boundary conditions. Therefore, the development of a robust and industry-compatible framework for accurate position and orientation estimation of flexible manipulators remains an essential challenge.

\subsection{Literature Review}

Numerous approaches have been investigated in the literature for the dynamic modeling of flexible manipulators~\cite{li2024advances}, ranging from infinite-dimensional formulations based on partial differential equations (PDEs), such as those derived from Hamilton's principle~\cite{he2020dynamical, liu2018dynamic}, to finite-dimensional methods including the finite element method (FEM)~\cite{kivila2021modeling}. The kinematics of the system, i.e., the position and orientation of each segment, are subsequently obtained by solving the governing dynamic equations, which remains a non-trivial task. For instance, non-homogeneous boundary conditions induced by gravitational effects complicate the derivation of accurate mode shapes~\cite{yaqubi2023semi}. Furthermore, solving these equations can be computationally intensive and may not consistently produce results that align with experimental observations~\cite{tahamipour2025computationally}. In addition, many control strategies rely on measurements of the flexible link’s end-effector position or elastic deflections~\cite{barjini2025deep}. Hence, the accurate estimation of the position and orientation of flexible manipulators remains a critical challenge, motivating further research in this domain.

Various methods have been proposed for obtaining the states of a flexible link, which generally fall into two categories: model-based state observers and sensor-based state measurements. Model-based observers estimate system states from limited measurements; however, their accuracy strongly depends on the fidelity of the underlying model, and parameter uncertainties can lead to significant estimation errors in practical applications~\cite{yaqubi2025nonlinear}. Moreover, even observer-based approaches require measurement inputs to function effectively~\cite{zhang2012observer}. 
Alternatively, sensor-based state measurement techniques for flexible manipulators primarily employ strain gauges~\cite{yin2017decomposed} or inertial measurement units (IMUs)~\cite{meng2023spring}. Strain gauges, however, often suffer from reduced reliability under extreme environmental conditions and may degrade during long-term dynamic operation. IMUs, while widely adopted, are susceptible to integration drift, and their accelerometers and gyroscopes are noise-sensitive, requiring advanced filtering and sensor fusion methods to achieve accurate state estimation~\cite{brossard2020denoising, li2022calib}.

\subsection{Motivation and Research Gap}

In the operation of flexible manipulators, elastic deformations and vibrations strongly affect the accuracy of end-effector motion control. This challenge is further amplified in heavy-duty flexible manipulators, where gravitational forces play a significant role~\cite{barjini2024deep}. Moreover, precise end-effector state information is essential for the implementation of PDE-based control strategies in flexible manipulators~\cite{barjini2025deep}. 
A major concern in kinematic estimation lies in both the cost and reliability of sensing technologies under diverse operating conditions. For instance, while laser trackers can provide highly accurate measurements, their high cost limits widespread industrial adoption. Similarly, although cost-efficient, strain gauges often suffer from reduced reliability in harsh environments (e.g., temperature extremes, high humidity, mechanical shock, and dust). 
Therefore, this research is motivated by the need to develop a methodology for the kinematic estimation of flexible manipulators that is accurate, reliable, and cost-effective, while remaining suitable for industrial deployment.

As highlighted in the literature, multi-IMU systems have been employed for end-effector kinematic estimation; however, the effect of filter parameters on estimation noise and temporal delay remains underexplored, and systematic parameter optimization is still an open research problem~\cite{vihonen2017joint, zhang2018angle}. Furthermore, although recent studies have applied IMUs for kinematic estimation in flexible manipulators, time-varying errors between the estimated and actual end-effector positions persist~\cite{tahamipour2024computationally, tahamipour2025computationally}, highlighting the need for improved refinement and correction strategies.

\subsection{Research Contributions and Structure of the Paper}

The primary contributions of this research are summarized as follows:
\begin{itemize}

\item Development of an AI-augmented IMU-based kinematic estimation framework applicable to rigid, flexible, and soft robots, enabling accurate end-effector position and orientation estimation in vertical motions.

\item Optimization of complementary filter parameters using Particle Swarm Optimization (PSO) to minimize sensor noise and delay effects in flexible links.

\item Compensation of residual position and orientation estimation errors in flexible links through a Radial Basis Function Neural Network (RBFNN), trained using laser tracker data as ground truth.

\end{itemize}

The remainder of this paper is organized as follows. Section~\ref{sec:kinematics} presents the IMU-based kinematic modeling approach. Section~\ref{sec:complementary} covers the complementary filtering method for sensor fusion, while Section~\ref{sec:filter} describes its PSO-based parameter optimization. Section~\ref{sec:surrogate} introduces an RBFNN-based model for residual error correction using ground-truth data. Experimental results are provided in Section~\ref{sec:experimental}, and conclusions and future directions are given in Section~\ref{sec:conclusions}.

\section{Kinematic Modeling Using IMU-Based Sensor Fusion}
\label{sec:kinematics}

This chapter introduces a kinematic modeling framework for flexible manipulators that employs multiple IMUs mounted along the flexible link. Using the proposed approach, the end-effector position and orientation are estimated from motion data recorded by the IMUs. The methodology is formulated to be general, allowing its application to manipulators with both rigid and flexible links.

\subsection{Modeling the Flexible Manipulator as a Chain of Rigid segments}
\label{sunsec:modeling}

\begin{figure}[h]
    \centering
    \includegraphics[width=1\linewidth]{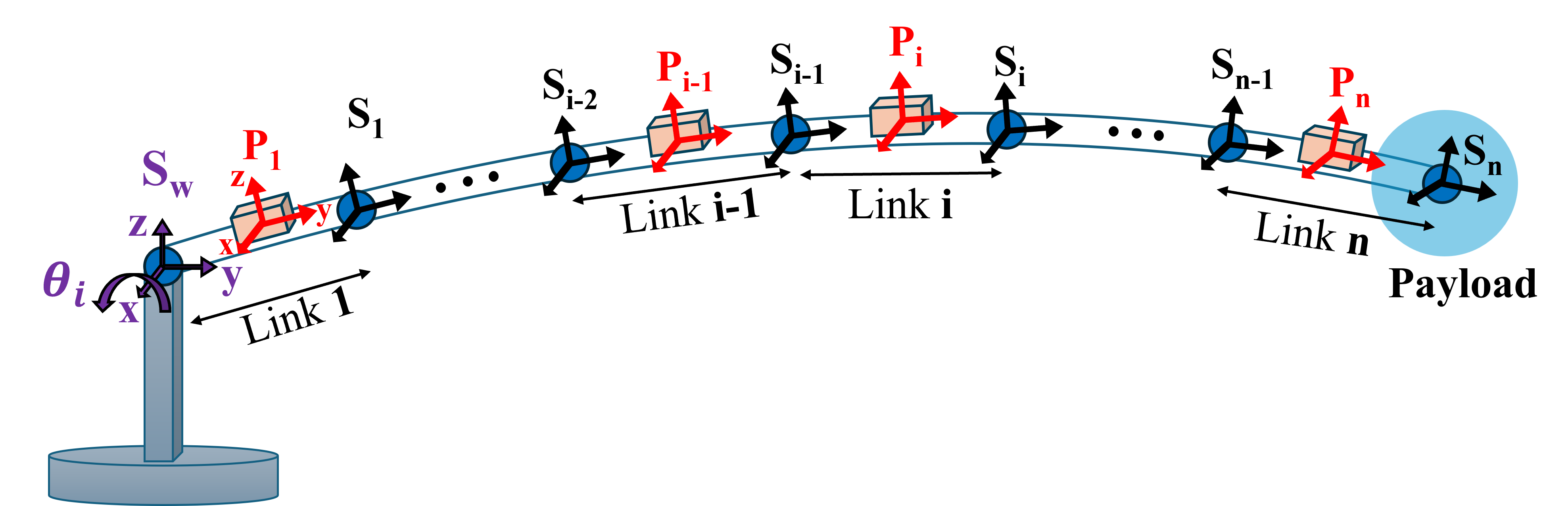}
    \caption{Flexible manipulator, modeled as a chain of rigid segments, with the corresponding frames.}
    \label{fig:flexible_link}
\end{figure}

Unlike rigid-link manipulators, where the relative joint angles alone are sufficient to determine the end-effector position and orientation, flexible-link manipulators are subject to elastic deformations and vibrations, making accurate kinematic estimation considerably more difficult. To address this challenge, the flexible link in this work is modeled as an open kinematic chain of $n$ rigid segments connected by revolute joints, as illustrated in Fig.~\ref{fig:flexible_link}.

As shown in Fig.~\ref{fig:flexible_link}, the Denavit--Hartenberg (DH) convention~\cite{sciavicco2012modelling} is adopted, where $S_i$ represents the body-fixed frame attached to the endpoint of the $i^\text{th}$ link. Furthermore, since IMUs serve as sensing elements, an auxiliary body-fixed frame $P_i$ is defined at the IMU location on the corresponding link.

\subsubsection*{Assumption 1}
\textit{The frames $S_i$ and $P_i$ are assumed to be aligned in orientation, such that they share the same angular velocity.}

\subsubsection*{Assumption 2}
\textit{The IMUs are rigidly mounted at fixed locations on the links. Consequently, the vector ${}^{i}_{i-1}\mathbf{r}_{P_i}$ (the position of frame $P_i$ with respect to frame $S_{i-1}$, expressed in frame $S_i$) and the vector ${}^{i-1}_{i-1}\mathbf{r}_{P_{i-1}}$ (the position of frame $P_{i-1}$ relative to frame $S_{i-1}$, expressed in frame $S_{i-1}$) remain constant over time. Therefore, their time derivatives are zero.}

Therefore, the end-effector position and orientation are modeled as follows:
\begin{equation}
\left\{
\begin{aligned}
y &= \sum_{i=1}^{n} l_i \cos\left( \sum_{j=1}^{i} \theta_j \right), \\
z &= \sum_{i=1}^{n} l_i \sin\left( \sum_{j=1}^{i} \theta_j \right), \\
\theta &= \sum_{i=1}^{n} \theta_i.
\end{aligned}
\right.
\label{eq:yztheta}
\end{equation}
Here, $n$ denotes the number of rigid segments assumed to model the flexible link, while $l_i$ and $\theta_i$ represent the length and joint angle of the $i^\text{th}$ segment, respectively. The relative joint angles are computed in Section~\ref{subsec:IMU} using data from the IMUs.

\subsection{IMU-Based Orientation and Motion Derivation}
\label{subsec:IMU}

\begin{figure}[h]
    \centering
    \includegraphics[width=1\linewidth]{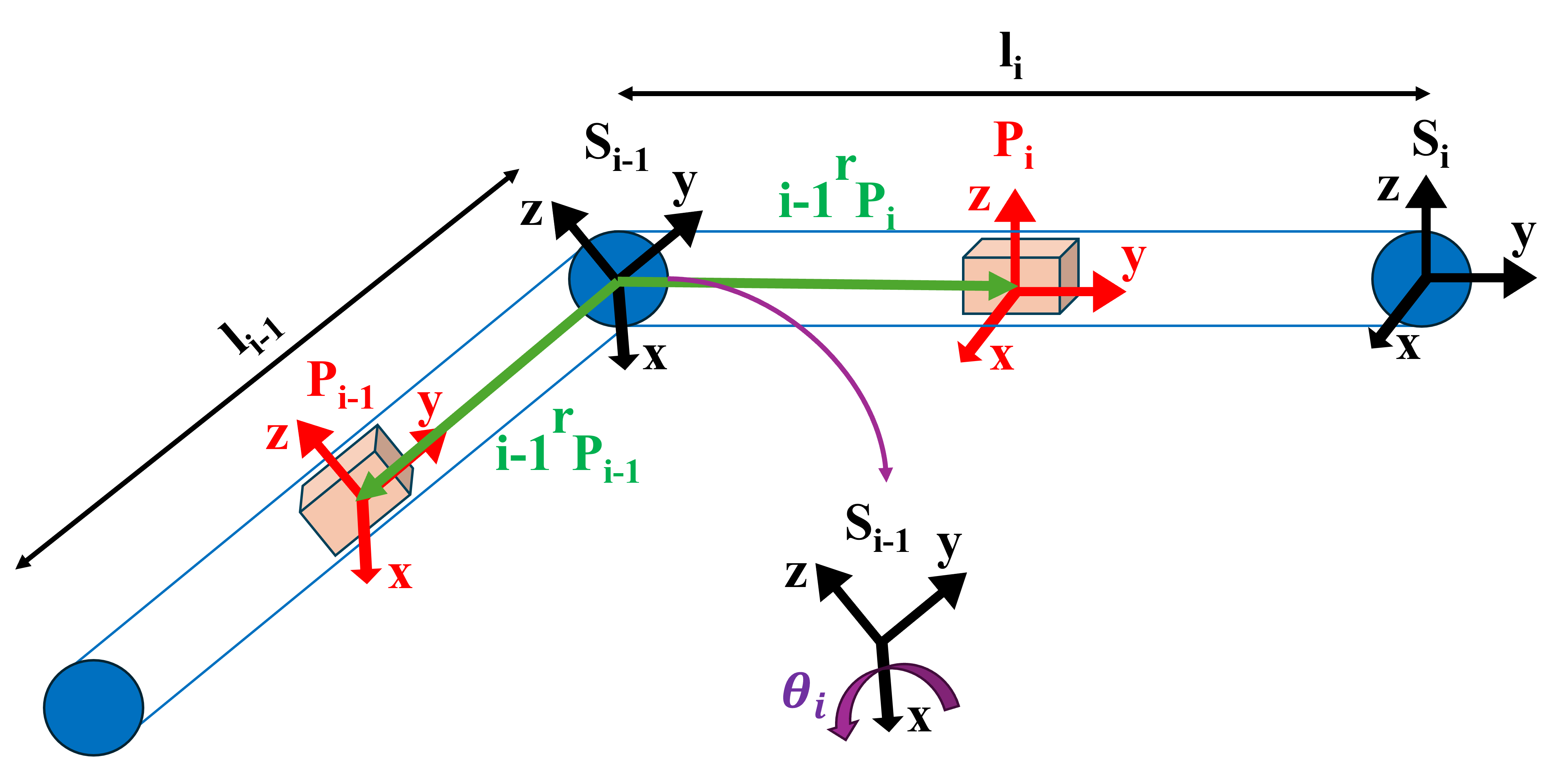}
    \caption{Schematic view of the link $i$ and $i-1$ and their IMUs.}
    \label{fig:links}
\end{figure}

Consider the $i^\text{th}$ link, as shown in Fig.~\ref{fig:links}, with its associated frame $S_i$, on which an IMU is mounted with a local frame denoted as $P_i$. Under the rigid-body assumption, the position vector of frame $P_i$ with respect to the inertial frame $W$, expressed in $W$ and denoted as ${}^{W}\mathbf{r}_{P_i}$, can be computed from the position vector of the preceding link frame $S_{i-1}$, denoted as ${}^{W}\mathbf{r}_{i-1}$, as follows:
\begin{equation}
    {}^{W}\mathbf{r}_{P_i} = {}^{W}\mathbf{r}_{i-1} + {}^{W}\mathbf{R}_{i} \, {}^{i}_{i-1}\mathbf{r}_{P_i},
\label{W_r_Pi}
\end{equation}
where ${}^{W}\mathbf{R}_{i} \in \mathbb{R}^{3 \times 3}$ denotes the rotation matrix from frame $S_i$ to the inertial frame $W$, and ${}^{i}_{i-1}\mathbf{r}_{P_i} \in \mathbb{R}^3$ represents the position vector of frame $P_i$ with respect to frame $S_{i-1}$, expressed in frame $S_i$. According to Assumption~2, the vector ${}^{i}_{i-1}\mathbf{r}_{P_i}$ remains constant, and its time derivatives are zero. Therefore, the velocity vector can be obtained by differentiating \eqref{W_r_Pi} as follows:
\begin{equation}
\begin{split}
    {}^{W}\dot{\mathbf{r}}_{P_i} &= {}^{W}\dot{\mathbf{r}}_{i-1} 
    + {}^{W}\mathbf{R}_{i} \, {}^{i}_{i-1}\dot{\mathbf{r}}_{P_i} 
    + \left({}^{W}\boldsymbol{\omega}_i \times\right) {}^{W}\mathbf{R}_{i} \, {}^{i}_{i-1}\mathbf{r}_{P_i} \\
    &= {}^{W}\dot{\mathbf{r}}_{i-1} 
    + \left({}^{W}\boldsymbol{\omega}_i \times\right) {}^{W}\mathbf{R}_{i} \, {}^{i}_{i-1}\mathbf{r}_{P_i},
\end{split}
\label{W_dr_Pi}
\end{equation}
where ${}^{W}\boldsymbol{\omega}_i \in \mathbb{R}^3$ denotes the angular velocity vector of frame $S_i$ (or equivalently $P_i$, according to Assumption~1), and $\left({}^{W}\boldsymbol{\omega}_i \times\right) \in \mathbb{R}^{3 \times 3}$ represents the corresponding skew-symmetric (cross-product) matrix. The acceleration vector is then obtained by differentiating \eqref{W_dr_Pi} as follows:
\begin{equation}
\begin{split}
    {}^{W}\ddot{\mathbf{r}}_{P_i} &= {}^{W}\ddot{\mathbf{r}}_{i-1} 
    + \left({}^{W}\dot{\boldsymbol{\omega}}_i \times\right) {}^{W}\mathbf{R}_{i} \, {}^{i}_{i-1}\mathbf{r}_{P_i} \\
    & \quad + \left({}^{W}\boldsymbol{\omega}_i \times\right) \left({}^{W}\boldsymbol{\omega}_i \times\right) {}^{W}\mathbf{R}_{i} \, {}^{i}_{i-1}\mathbf{r}_{P_i}\\ 
    & \quad + \left({}^{W}\boldsymbol{\omega}_i \times\right) {}^{W}\mathbf{R}_{i} \, {}^{i}_{i-1}\dot{\mathbf{r}}_{P_i}\\
    &= {}^{W}\ddot{\mathbf{r}}_{i-1} + \left({}^{W}\dot{\boldsymbol{\omega}}_i \times\right) {}^{W}\mathbf{R}_{i} \, {}^{i}_{i-1}\mathbf{r}_{P_i}\\ 
    & \quad + \left({}^{W}\boldsymbol{\omega}_i \times\right) \left({}^{W}\boldsymbol{\omega}_i \times\right) {}^{W}\mathbf{R}_{i} \, {}^{i}_{i-1}\mathbf{r}_{P_i}.
\end{split}
\label{W_ddr_Pi}
\end{equation}

It should be noted that ${}^{W}\ddot{\mathbf{r}}_{P_i}$ represents the acceleration of frame $P_i$ with respect to the inertial frame $W$. However, the accelerometers in the IMUs measure acceleration in their local coordinate frame. Therefore, \eqref{W_ddr_Pi} must be multiplied by ${}^{W}\mathbf{R}_{i}^T$ to express the acceleration in the local frame. To begin, the following equation is considered:
\begin{equation}
    \left({}^{W}\boldsymbol{\omega}_i \times\right) = {}^{W}\mathbf{R}_{i} \left({}^{i}\boldsymbol{\omega}_i \times\right) {}^{W}\mathbf{R}_{i}^T,
\label{W_omega_i}
\end{equation}
then, multiplying \eqref{W_ddr_Pi} by ${}^{W}\mathbf{R}_{i}^T$ yields:
\begin{equation}
    {}^{i}\ddot{\mathbf{r}}_{P_i} = {}^{i}\ddot{\mathbf{r}}_{i-1} + \left({}^{i}\dot{\boldsymbol{\omega}}_i \times\right) {}^{i}_{i-1}\mathbf{r}_{P_i} + \left({}^{i}\boldsymbol{\omega}_i \times\right) \left({}^{i}\boldsymbol{\omega}_i \times\right) {}^{i}_{i-1}\mathbf{r}_{P_i},
\label{i_ddr_Pi}
\end{equation}
where ${}^{i}\ddot{\mathbf{r}}_{P_i}$ denotes the instantaneous acceleration of the IMU frame $P_i$, expressed in its local coordinate frame.

The same procedure can now be applied to the ${(i\!-\!1)}^\text{th}$ link, as shown in Fig.~\ref{fig:links}. Since the primary objective of this section is to compute the joint angle $\theta_i$, we start from frame $S_{i-1}$ to determine the position vector of frame $P_{i-1}$ as follows:
\begin{equation}
    {}^{W}\mathbf{r}_{P_{i-1}} = {}^{W}\mathbf{r}_{i-1} + {}^{W}\mathbf{R}_{i-1} \, {}^{i-1}_{i-1}\mathbf{r}_{P_{i-1}},
\label{W_r_Pi-1}
\end{equation}
where ${}^{i-1}_{i-1}\mathbf{r}_{P_{i-1}} \in \mathbb{R}^3$ denotes the position vector of frame $P_{i-1}$ with respect to frame $S_{i-1}$, expressed in the coordinate frame $S_{i-1}$. According to Assumption~2, since ${}^{i-1}_{i-1}\mathbf{r}_{P_{i-1}}$ is constant, it follows that:
\begin{equation}
\begin{split}
    {}^{W}\dot{\mathbf{r}}_{P_i} &= {}^{W}\dot{\mathbf{r}}_{i-1} 
    + {}^{W}\mathbf{R}_{i-1} \, {}^{i-1}_{i-1}\dot{\mathbf{r}}_{P_{i-1}}\\ 
    & \quad + \left({}^{W}\boldsymbol{\omega}_{i-1} \times\right) {}^{W}\mathbf{R}_{i-1} \, {}^{i-1}_{i-1}\mathbf{r}_{P_{i-1}} \\
    &= {}^{W}\dot{\mathbf{r}}_{i-1} 
    + \left({}^{W}\boldsymbol{\omega}_{i-1} \times\right) {}^{W}\mathbf{R}_{i-1} \, {}^{i-1}_{i-1}\mathbf{r}_{P_{i-1}}.
\end{split}
\label{W_dr_Pi-1}
\end{equation}
By differentiating it once more, we obtain:

\begin{equation}
\begin{split}
    {}^{W}\ddot{\mathbf{r}}_{P_{i-1}} &= {}^{W}\ddot{\mathbf{r}}_{i-1} 
    + \left({}^{W}\dot{\boldsymbol{\omega}}_{i-1} \times\right) {}^{W}\mathbf{R}_{i-1} \, {}^{i-1}_{i-1}\mathbf{r}_{P_{i-1}} \\
    & \quad + \left({}^{W}\boldsymbol{\omega}_{i-1} \times\right) \left({}^{W}\boldsymbol{\omega}_{i-1} \times\right) {}^{W}\mathbf{R}_{i-1} \, {}^{i-1}_{i-1}\mathbf{r}_{P_{i-1}}\\ 
    & \quad + \left({}^{W}\boldsymbol{\omega}_{i-1} \times\right) {}^{W}\mathbf{R}_{i-1} \, {}^{i-1}_{i-1}\dot{\mathbf{r}}_{P_{i-1}}\\
    &= {}^{W}\ddot{\mathbf{r}}_{i-1} + \left({}^{W}\dot{\boldsymbol{\omega}}_{i-1} \times\right) {}^{W}\mathbf{R}_{i-1} \, {}^{i-1}_{i-1}\mathbf{r}_{P_{i-1}}\\ 
    & \quad + \left({}^{W}\boldsymbol{\omega}_{i-1} \times\right) \left({}^{W}\boldsymbol{\omega}_{i-1} \times\right) {}^{W}\mathbf{R}_{i-1} \, {}^{i-1}_{i-1}\mathbf{r}_{P_{i-1}}.
\end{split}
\label{W_ddr_Pi-1}
\end{equation}

Similarly to the previous case, consider the following relation:
\begin{equation}
    \left({}^{W}\boldsymbol{\omega}_{i-1} \times\right) = {}^{W}\mathbf{R}_{i-1} \left({}^{i-1}\boldsymbol{\omega}_{i-1} \times\right) {}^{W}\mathbf{R}_{i-1}^T,
\label{W_omega_i-1}
\end{equation}
and multiplying \eqref{W_ddr_Pi-1} by ${}^{W}\mathbf{R}_{i-1}^T$ to express the instantaneous acceleration of the IMU frame $P_{i-1}$ in its local coordinate frame yields:
\begin{equation}
    \begin{split}
    {}^{i-1}\ddot{\mathbf{r}}_{P_{i-1}} &= {}^{i-1}\ddot{\mathbf{r}}_{i-1} + \left({}^{i-1}\dot{\boldsymbol{\omega}}_{i-1} \times\right) {}^{i-1}_{i-1}\mathbf{r}_{P_{i-1}}\\
    & \quad + \left({}^{i-1}\boldsymbol{\omega}_{i-1} \times\right) \left({}^{i-1}\boldsymbol{\omega}_{i-1} \times\right) {}^{i-1}_{i-1}\mathbf{r}_{P_{i-1}}.
    \end{split}
\label{i-1_ddr_Pi-1}
\end{equation}

Equations \eqref{i_ddr_Pi} and \eqref{i-1_ddr_Pi-1} are employed in Section~\ref{subsec:gravity} to derive the joint angle $\theta_i$, representing the relative rotation between two consecutive segments. By combining accelerometer and gyroscope measurements from adjacent IMUs with known geometric parameters, $\theta_i$ can be estimated accurately.

\subsection{Gravity-Referenced Joint Angle Estimation}
\label{subsec:gravity}

To derive the joint angle $\theta_i$, measurements from the ${i}^\text{th}$ and ${(i\!-\!1)}^\text{th}$ IMUs, located at frames $P_i$ and $P_{i-1}$, respectively, are required. The relationship between these measurements is expressed by the following equation:
\begin{equation}
    {}^{i-1}\ddot{\mathbf{r}}_{P_{i-1}} = {}^{i-1}\mathbf{R}_{i} \, {}^{i}\ddot{\mathbf{r}}_{P_{i-1}} = {}^{i-1}\mathbf{R}_{i} \, \left( {}^{i}\ddot{\mathbf{r}}_{P_i} - {}^{i}_{P_{i-1}}\ddot{\mathbf{r}}_{P_i} \right),
\label{i-1_ddotr_Pi-1}
\end{equation}
where ${}^{i-1}\ddot{\mathbf{r}}_{P_{i-1}}$ and ${}^{i}\ddot{\mathbf{r}}_{P_i}$ denote the instantaneous acceleration vectors of the IMU frames $P_{i-1}$ and $P_i$, expressed in their respective local coordinate frames. The remaining terms to be determined are ${}^{i-1}\mathbf{R}_{i}$, the rotation matrix representing the orientation of frame $S_i$ with respect to frame $S_{i-1}$, and ${}^{i}_{P_{i-1}}\ddot{\mathbf{r}}_{P_i}$, the acceleration of frame $P_i$ relative to frame $P_{i-1}$, expressed in frame $S_i$. By substituting from \eqref{i_ddr_Pi} and \eqref{i-1_ddr_Pi-1}, ${}^{i}_{P_{i-1}}\ddot{\mathbf{r}}_{P_i}$ can be obtained as:
\begin{equation}
\begin{split}
    {}^{i}_{P_{i-1}}\ddot{\mathbf{r}}_{P_i} &= {}^{i}\ddot{\mathbf{r}}_{P_i} - {}^{i}\ddot{\mathbf{r}}_{P_{i-1}} = {}^{i}\ddot{\mathbf{r}}_{P_i} - {}^{i-1}\mathbf{R}_{i}^T \, {}^{i-1}\ddot{\mathbf{r}}_{P_{i-1}}\\ 
    &= \left({}^{i}\dot{\boldsymbol{\omega}}_i \times\right) {}^{i}_{i-1}\mathbf{r}_{P_i} - \left({}^{i-1}\dot{\boldsymbol{\omega}}_{i-1} \times\right) {}^{i-1}_{i-1}\mathbf{r}_{P_{i-1}}\\
    & \quad + \left({}^{i}\boldsymbol{\omega}_i \times\right) \left({}^{i}\boldsymbol{\omega}_i \times\right) {}^{i}_{i-1}\mathbf{r}_{P_i}\\
    & \quad - \left({}^{i-1}\boldsymbol{\omega}_{i-1} \times\right) \left({}^{i-1}\boldsymbol{\omega}_{i-1} \times\right) {}^{i-1}_{i-1}\mathbf{r}_{P_{i-1}}.
\label{Pi-1_ddr_Pi}
\end{split}
\end{equation}
Assuming that the joints between the rigid segments are revolute, as shown in Fig.~\ref{fig:links}, the rotation matrix takes the following form:
\begin{equation}
{}^{i-1}\mathbf{R}_{i} = 
\begin{bmatrix}
1 & 0 & 0 \\
0 & \cos\theta_i & -\sin\theta_i \\
0 & \sin\theta_i & \cos\theta_i
\end{bmatrix}.
\label{Rotation_Matrix}
\end{equation}
By considering \eqref{i-1_ddotr_Pi-1} and substituting \eqref{Pi-1_ddr_Pi} and \eqref{Rotation_Matrix}, the following expression can be derived:
\begin{equation}
\begin{bmatrix}
{}^{i-1}\ddot{\mathbf{r}}^x_{P_{i-1}} \\
{}^{i-1}\ddot{\mathbf{r}}^y_{P_{i-1}} \\
{}^{i-1}\ddot{\mathbf{r}}^z_{P_{i-1}}
\end{bmatrix}
= 
\begin{bmatrix}
1 & 0 & 0 \\
0 & \cos\theta_i & -\sin\theta_i \\
0 & \sin\theta_i & \cos\theta_i
\end{bmatrix}
\begin{bmatrix}
{}^{i}\ddot{\mathbf{r}}^x_{P_i} - {}^{i}_{P_{i-1}}\ddot{\mathbf{r}}^x_{P_i} \\
{}^{i}\ddot{\mathbf{r}}^y_{P_i} - {}^{i}_{P_{i-1}}\ddot{\mathbf{r}}^y_{P_i} \\
{}^{i}\ddot{\mathbf{r}}^z_{P_i} - {}^{i}_{P_{i-1}}\ddot{\mathbf{r}}^z_{P_i}
\end{bmatrix}.
\label{eq:main1}
\end{equation}
Expanding the matrix multiplication, the $y$ and $z$ components of the acceleration in frame $S_{i-1}$ can be explicitly written as:
\begin{equation}
\left\{
\begin{aligned}
{}^{i-1}\ddot{\mathbf{r}}^y_{P_{i-1}} &= \cos\theta_i \left( {}^{i}\ddot{\mathbf{r}}^y_{P_i} - {}^{i}_{P_{i-1}}\ddot{\mathbf{r}}^y_{P_i} \right)\\
&  - \sin\theta_i \left( {}^{i}\ddot{\mathbf{r}}^z_{P_i} - {}^{i}_{P_{i-1}}\ddot{\mathbf{r}}^z_{P_i} \right), \\
{}^{i-1}\ddot{\mathbf{r}}^z_{P_{i-1}} &= \sin\theta_i \left( {}^{i}\ddot{\mathbf{r}}^y_{P_i} - {}^{i}_{P_{i-1}}\ddot{\mathbf{r}}^y_{P_i} \right)\\
&  + \cos\theta_i \left( {}^{i}\ddot{\mathbf{r}}^z_{P_i} - {}^{i}_{P_{i-1}}\ddot{\mathbf{r}}^z_{P_i} \right).
\end{aligned}
\right.
\label{eq:main2}
\end{equation}
Consequently, the joint angle $\theta_i$ can be computed from \eqref{eq:main2} as:
\begin{equation}
\left\{
\begin{aligned}
    \theta_i &= \tan^{-1}\left( \frac{A_{i}}{B_{i}} \right),\\
    A_{i} &= {}^{i-1}\ddot{\mathbf{r}}^z_{P_{i-1}} \left( {}^{i}\ddot{\mathbf{r}}^y_{P_i} - {}^{i}_{P_{i-1}}\ddot{\mathbf{r}}^y_{P_i} \right)\\
    & - {}^{i-1}\ddot{\mathbf{r}}^y_{P_{i-1}} \left( {}^{i}\ddot{\mathbf{r}}^z_{P_i} - {}^{i}_{P_{i-1}}\ddot{\mathbf{r}}^z_{P_i} \right),\\
    B_{i} &= {}^{i-1}\ddot{\mathbf{r}}^y_{P_{i-1}} \left( {}^{i}\ddot{\mathbf{r}}^y_{P_i} - {}^{i}_{P_{i-1}}\ddot{\mathbf{r}}^y_{P_i} \right)\\
    & + {}^{i-1}\ddot{\mathbf{r}}^z_{P_{i-1}} \left( {}^{i}\ddot{\mathbf{r}}^z_{P_i} - {}^{i}_{P_{i-1}}\ddot{\mathbf{r}}^z_{P_i} \right).
\end{aligned}
\right.
\label{eq:theta}
\end{equation}

\subsection{Joint Angular Velocity Estimation}
\label{subsec:joint}

To estimate the joint angular velocity $\dot{\theta}_i$, we first define the relative angular velocity between the $i^\text{th}$ and $(i-1)^\text{th}$ links. This relative angular velocity, expressed in the coordinate frame of the $i^\text{th}$ link, is given by:
\begin{equation}
    {}^{i}_{i-1}\boldsymbol{\omega}_{i} = {}^{i}\boldsymbol{\omega}_{i} - {}^{i}\boldsymbol{\omega}_{i-1} = {}^{i}\boldsymbol{\omega}_{i} - {}^{i-1}\mathbf{R}^T_{i} \, {}^{i-1}\boldsymbol{\omega}_{i-1},
\label{eq:i-1_omega_i}
\end{equation}
where ${}^{i}\boldsymbol{\omega}_{i}$ and ${}^{i-1}\boldsymbol{\omega}_{i-1}$ denote the angular velocity vectors at frames $P_i$ and $P_{i-1}$, respectively, and ${}^{i-1}\mathbf{R}_{i}$ represents the rotation matrix from frame $S_i$ to frame $S_{i-1}$.

Since the joint is assumed to be revolute about the $x$-axis of frame $S_i$, the joint angular velocity can be directly obtained from the $x$-component of the relative angular velocity vector:
\begin{equation}
    \dot{\theta}_i = {}^{i}_{i-1}\boldsymbol{\omega}^x_{i}.
\label{eq:theta_dot}
\end{equation}

\section{Complementary Filtering}
\label{sec:complementary}

\subsection{Impact of Noise, Drift, and Delay on IMU-Based Estimation}
\label{subsec:impact}

Theoretically, Section~\ref{sec:kinematics} introduced a framework for estimating the end-effector's position and orientation in rigid, flexible, and soft robots. However, in practice, IMU measurements are susceptible to noise and drift, which can degrade the accuracy of the kinematic estimation. Therefore, filtering these measurements becomes essential. 

To account for noise and drift effects, the gyroscope and accelerometer measurements are modeled as follows~\cite{forster2016manifold}:
\begin{equation}
\left\{
\begin{aligned}
{}^{i}\boldsymbol{\omega_{i_{\text{meas}}}} &= {}^{i}\boldsymbol{\omega}_i + {}^{i} \mathbf{b}_{\omega_i} + {}^{i} \mathbf{n}_{\omega_i},\\
{}^{i}\mathbf{a}_{{P_i}_{\text{meas}}} &= {}^{i}\ddot{\mathbf{r}}_{P_i} - {}^{W}\mathbf{R}_{i}^T \, \mathbf{g}  + {}^{i}\mathbf{b}_{a_i} + {}^{i}\mathbf{n}_{a_i},
\end{aligned}
\right.
\label{eq:measurements}
\end{equation}
where ${}^{i}\boldsymbol{\omega}_{i_{\text{meas}}}$ and ${}^{i}\mathbf{a}_{{P_i}_{\text{meas}}}$ denote the angular velocity and linear acceleration measurements of the $i^\text{th}$ IMU, respectively; ${}^{i} \mathbf{b}_{\omega_i}$ and ${}^{i}\mathbf{b}_{a_i}$ represent the corresponding measurement biases, and ${}^{i} \mathbf{n}_{\omega_i}$ and ${}^{i}\mathbf{n}_{a_i}$ denote the measurement noise components. The vector $\mathbf{g} = [0,\, 0,\,-9.81]^\top$ represents the gravitational acceleration acting on the IMUs, expressed in the inertial frame.

Due to the presence of bias and noise in the IMU measurements, the estimated joint angle and angular velocity presented in~\eqref{eq:theta} and~\eqref{eq:theta_dot} should be expressed as:
\begin{equation}
\left\{
\begin{aligned}
    \theta_{{i}_{\text{meas}}} &= \tan^{-1}\left( \frac{A_{{i}_{\text{meas}}}}{B_{{i}_{\text{meas}}}} \right),\\
    A_{{i}_{\text{meas}}} &= {}^{i-1}\mathbf{a}^z_{{P_{i-1}}_{\text{meas}}} \left( {}^{i}\mathbf{a}^y_{{P_{i}}_{\text{meas}}} - {}^{i}_{P_{i-1}}\ddot{\mathbf{r}}^y_{{P_i}_{\text{meas}}} \right)\\
    & - {}^{i-1}\mathbf{a}^y_{{P_{i-1}}_{\text{meas}}} \left( {}^{i}\mathbf{a}^z_{{P_{i}}_{\text{meas}}} - {}^{i}_{P_{i-1}}\ddot{\mathbf{r}}^z_{{P_i}_{\text{meas}}} \right),\\
    B_{{i}_{\text{meas}}} &= {}^{i-1}\mathbf{a}^y_{{P_{i-1}}_{\text{meas}}} \left( {}^{i}\mathbf{a}^y_{{P_{i}}_{\text{meas}}} - {}^{i}_{P_{i-1}}\ddot{\mathbf{r}}^y_{{P_i}_{\text{meas}}} \right)\\
    & + {}^{i-1}\mathbf{a}^z_{{P_{i-1}}_{\text{meas}}} \left( {}^{i}\mathbf{a}^z_{{P_{i}}_{\text{meas}}} - {}^{i}_{P_{i-1}}\ddot{\mathbf{r}}^z_{{P_i}_{\text{meas}}} \right),
\end{aligned}
\right.
\label{eq:theta_meas}
\end{equation}

\begin{equation}
\left\{
\begin{aligned}
    \dot{\theta}_{i_{\text{meas}}} &= {}^{i}_{i-1}\boldsymbol{\omega}^x_{i_{\text{meas}}},\\
    {}^{i}_{i-1}\boldsymbol{\omega}_{i_{\text{meas}}} &= {}^{i}\boldsymbol{\omega}_{i_{\text{meas}}} - {}^{i-1}\mathbf{R}^T_{i} \, {}^{i-1}\boldsymbol{\omega}_{{i-1}_{\text{meas}}}.
\end{aligned}
\right.
\label{eq:theta_dot_meas}
\end{equation}

\subsection{Theory of Complementary Filtering}
\label{subsec:theory}

As joint angle and angular velocity are both available from the IMU-based sensing system, they can be fused to yield a more accurate and robust joint angle estimate using a complementary filtering approach~\cite{higgins2007comparison, wu2016fast}. In this context, the measured joint angle $\theta_{{i}_{\text{meas}}}$ typically contains high-frequency noise, while the angular velocity $\dot{\theta}_{i_{\text{meas}}}$, when integrated over time, accumulates low-frequency drift. To mitigate both effects, the following complementary filter is applied to estimate a joint angle that is both noise-reduced and drift-compensated~\cite{vihonen2017joint}:
\begin{equation}
\begin{bmatrix}
\dot{\hat{\theta}}_i \\
\dot{\hat{b}}_i
\end{bmatrix}
=
\begin{bmatrix}
0 & 1 \\
0 & 0
\end{bmatrix}
\begin{bmatrix}
\hat{\theta}_i \\
\hat{b}_i
\end{bmatrix}
+
\begin{bmatrix}
k_P \\
k_I
\end{bmatrix}
\left( \theta_{i_\text{meas}} - \hat{\theta}_i \right)
+
\begin{bmatrix}
1 \\
0
\end{bmatrix}
\dot{\theta}_{i_\text{meas}},
\label{eq:filter}
\end{equation}
where $\hat{\theta}_i$ denotes the filtered joint angle, $K_P$ and $K_I$ are tunable filter gains, and $\theta_{i_{\text{meas}}}$ and $\dot{\theta}_{i_{\text{meas}}}$ represent the measured joint angle and angular velocity, respectively.

\begin{figure}[h]
    \centering
    \includegraphics[width=1\linewidth]{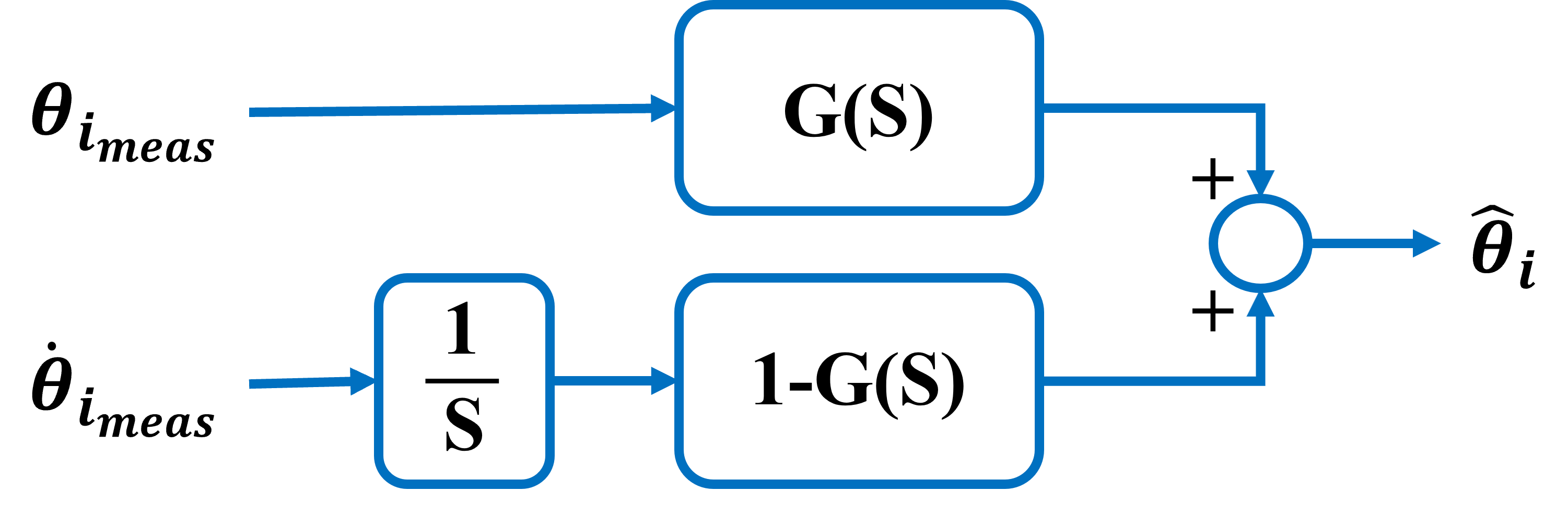}
    \caption{Block diagram of the complimentary filtering}
    \label{fig:complementary_filtering}
\end{figure}

In fact, the proposed complementary filter in \eqref{eq:filter} fuses joint angle and angular velocity measurements by applying a low-pass filter to $\theta_{i_{\text{meas}}}$ and a high-pass filter to $\int_0^t \dot{\theta}_{i_{\text{meas}}} \, dt$. To illustrate this behavior, we first expand the dynamic equations of the filter:
\begin{equation}
\begin{cases}
\dot{\hat{\theta}}_i = \hat{b}_i + k_P \left( \theta_{i_\text{meas}} - \hat{\theta}_i \right) + \dot{\theta}_{i_\text{meas}}, \\
\dot{\hat{b}}_i = k_I \left( \theta_{i_\text{meas}} - \hat{\theta}_i \right).
\end{cases}
\label{1}
\end{equation}
Differentiating the first equation with respect to time and substituting $\dot{\hat{b}}_i$ yields:
\begin{equation}
    \ddot{\hat{\theta}}_i = k_I \left( \theta_{i_\text{meas}} - \hat{\theta}_i \right) + k_P \left( \dot{\theta}_{i_\text{meas}} - \dot{\hat{\theta}}_i \right) + \ddot{\theta}_{i_\text{meas}}.
\label{2}
\end{equation}
Rearranging the terms leads to a second-order linear differential equation:
\begin{equation}
    \ddot{\hat{\theta}}_i + k_P \dot{\hat{\theta}}_i + k_I \hat{\theta}_i = \ddot{\theta}_{i_\text{meas}} + k_P \dot{\theta}_{i_\text{meas}} + k_I \theta_{i_\text{meas}}.
\label{3}
\end{equation}
Taking the Laplace transform of both sides, results in:
\begin{equation}
    \left( S^2 + K_P S + K_I \right) \hat{\theta}_i = \left( S \right) \dot{\theta}_{i_\text{meas}} + \left( K_P S + k_I \right) \theta_{i_\text{meas}}.
\label{4}
\end{equation}
Solving for $\hat{\theta}_i(S)$ yields:
\begin{equation}
\begin{split}
    \hat{\theta}_i &= \left( \frac{S^2}{S^2 + K_P S + K_I} \right) \frac{\dot{\theta}_{i_\text{meas}}}{S} \\
    &\quad + \left( \frac{K_P S + k_I}{S^2 + K_P S + K_I} \right) \theta_{i_\text{meas}} \\
    &= \left( 1 - G(S) \right) \frac{\dot{\theta}_{i_\text{meas}}}{S} + \left( G(S) \right) \theta_{i_\text{meas}},
\end{split}
\label{5}
\end{equation}
where $G(S)$ acts as a second-order low-pass filter, while $1 - G(S)$ serves as its complementary high-pass counterpart, as demonstrated in Fig.~\ref{fig:complementary_filtering}.

\subsection{Kinematic Modeling of the Flexible Manipulator Using the Filtered Joint Angles}

After computing the filtered joint angles $\hat{\theta}_i$, the corresponding end-effector pose of the flexible robot, comprising position and orientation, can be estimated as follows:
\begin{equation}
\left\{
\begin{aligned}
\hat{y} &= \sum_{i=1}^{n} l_i \cos\left( \sum_{j=1}^{i} \hat{\theta}_j \right), \\
\hat{z} &= \sum_{i=1}^{n} l_i \sin\left( \sum_{j=1}^{i} \hat{\theta}_j \right), \\
\hat{\theta} &= \sum_{i=1}^{n} \hat{\theta}_i.
\end{aligned}
\right.
\label{eq:yzthetahat}
\end{equation}
Here, $\hat{y}$ and $\hat{z}$ represent the estimated Cartesian coordinates of the end-effector, and $\hat{\theta}$ denotes its orientation, all derived from the filtered joint angles.

\section{Filter Parameter Optimization Using PSO}
\label{sec:filter}

The complementary filter described in Section~\ref{sec:complementary} relies on two gain parameters, $K_P$ and $K_I$. While prior studies have overlooked the fine-tuning of these gains~\cite{vihonen2017joint, tahamipour2024computationally, tahamipour2025computationally}, they have a significant impact on noise levels, bias, and delay between the estimated end-effector states ($\hat{y}$, $\hat{z}$, $\hat{\theta}$) and their corresponding ground truth values ($y_\text{gt}$, $z_\text{gt}$, $\theta_\text{gt}$). As a result, careful optimization of these gains is critical to achieving accurate kinematic estimation. 

To tune the complementary filter parameters, this work employs Particle Swarm Optimization (PSO)~\cite{kennedy1995particle}, which iteratively searches for the optimal gain values. Ground truth data for this study was obtained using the Leica Absolute Laser Tracker AT960-LR, a state-of-the-art portable metrology system designed for high-precision 3D tracking. The cost function is defined as:
\begin{equation}
J = - \left( \text{delay\_penalty} + \text{noise\_penalty} + \text{error\_penalty} \right),
\label{eq:cost}
\end{equation}
where $\text{delay\_penalty}$, $\text{noise\_penalty}$, and $\text{error\_penalty}$ represent penalization terms for temporal delay, signal noise, and estimation error, respectively.

The delay penalty quantifies the temporal misalignment between the estimated signal and the ground truth signal. This is computed based on the time difference between the peak locations of the two signals. Let $\text{loc}_{\text{est}}$ and $\text{loc}_{\text{gt}}$ denote the sample indices at which $\hat{z}$ and $z_\text{gt}$ reach their respective peaks. The delay penalty is then defined as:
\begin{equation}
\text{delay\_penalty} = K_{\text{delay}} \int_0^{t_{\text{end}}} T_s \cdot \left| \text{loc}_{\text{est}} - \text{loc}_{\text{gt}} \right| dt,
\label{eq:delay_penalty}
\end{equation}
where $T_s$ is the sampling time, and $K_{\text{delay}}$ is a positive gain. In this work, we use $T_s = 0.001$ seconds, and the peak locations are obtained using a buffer of historical values in Simulink.

The noise penalty is intended to quantify high-frequency fluctuations in the estimated position signal, which typically indicate measurement or estimation noise. To capture this, we first compute the discrete-time derivative of the estimated position:
\begin{equation}
\Delta P_{\text{est}}[k] = P_{\text{est}}[k] - P_{\text{est}}[k-1],
\label{eq:discrete_derivative}
\end{equation}
where $P_{\text{est}}[k] = \sqrt{ \hat{y}[k]^2 + \hat{z}[k]^2 }$ and $k$ is the discrete time index. This derivative is then passed through a high-pass filter to isolate high-frequency components. The noise penalty is defined as:
\begin{equation}
\text{noise\_penalty} = K_{\text{noise}} \int_0^{t_{\text{end}}}  \text{HPF} \left\{ \Delta P_{\text{est}}[k] \right\} dt ,
\label{eq:noise_penalty}
\end{equation}
where $\text{HPF}\{\cdot\}$ denotes the high-pass filter operation, and $K_{\text{noise}}$ is a positive gain.

The error penalty quantifies the Euclidean distance between the estimated and ground truth positions in the 2D plane. First, the error is defined as:
\begin{equation}
\text{error}(t) = \sqrt{ \left( \hat{y}(t) - y_{\text{gt}}(t) \right)^2 + \left( \hat{z}(t) - z_{\text{gt}}(t) \right)^2 },
\label{eq:error_penalty1}
\end{equation}
where $\hat{y}(t)$ and $\hat{z}(t)$ are the estimated positions, and $y_{\text{gt}}(t)$ and $z_{\text{gt}}(t)$ are the corresponding ground truth values at time step $t$. Then the error penalty is calculated as:
\begin{equation}
\text{error\_penalty} = K_{\text{error}} \int_0^{t_{\text{end}}} \text{error}(t) dt,
\label{eq:error_penalty}
\end{equation}
where $K_{\text{error}}$ is a positive value.

Finally, the gain parameters $K_P$ and $K_I$ are optimized using PSO, by minimizing the cost function:
\begin{equation}
\min_{K_P, K_I} J(K_P, K_I),
\label{eq:gain_optimization}
\end{equation}
subject to the following constraints:
\begin{equation}
0.1 \leq K_P \leq 100, \quad 0.01 \leq K_I \leq 10.
\end{equation}
With $K_{\text{delay}} = 200$, $K_{\text{noise}} = 5 \times 10^6$ and $K_{\text{error}} = 1$, the optimization produces the optimal gains as:
\begin{equation}
K_P = 7.0407, \quad K_I = 0.1984.
\end{equation}
These values were selected to balance noise and delay robustness, and low tracking error in static and dynamic conditions.

\section{AI-Augmented Modeling for Residual Correction Using RBFNN}
\label{sec:surrogate}

Although the filter parameters have been fine-tuned in Section~\ref{sec:filter}, a discrepancy still exists between the true position and orientation of the end-effector on the flexible link, denoted by ($y_\text{gt}$, $z_\text{gt}$, $\theta_\text{gt}$) from the ground-truth data, and the corresponding estimated values ($\hat{y}$, $\hat{z}$, $\hat{\theta}_i$) produced by the proposed multi-IMU sensor fusion system. This discrepancy arises due to model uncertainty, as the flexible link is approximated as a series of rigid segments. To address this limitation, this paper introduces the use of a radial basis function neural network (RBFNN) to learn and compensate for the residual error between the estimated and actual end-effector positions and orientations in flexible manipulators.

First, let the estimated state vector be defined as $\hat{\mathbf{X}} = [\hat{y}, \hat{z}, \hat{\theta}]$, the ground-truth state vector as $\mathbf{X}_\text{gt} = [y_\text{gt}, z_\text{gt}, \theta_\text{gt}]$, and the residual error vector as $\Delta \mathbf{X} = \mathbf{X}_\text{gt} - \hat{\mathbf{X}}$. The goal is to train an RBFNN to model the mapping:
\begin{equation}
    f_\text{RBF}: \hat{\mathbf{X}} \mapsto \Delta \mathbf{X}.
\label{eq:RBFNN1}
\end{equation}
The structure of the RBFNN is given by:
\begin{equation}
\left\{
\begin{aligned}
f_\text{RBF}(\hat{\mathbf{X}}) &= \sum_{i=1}^{n} \mathbf{w}_i \cdot \phi_i(\hat{\mathbf{X}}),\\
\phi_i(\hat{\mathbf{X}}) &= \exp\left( -\frac{\|\hat{\mathbf{X}} - \mathbf{c}_i\|^2}{2\sigma^2} \right),
\end{aligned}
\right.
\label{eq:RBFNN2}
\end{equation}
where $\mathbf{w}_i \in \mathbb{R}^3$ are the output weights associated with the $i$-th neuron, $\mathbf{c}_i \in \mathbb{R}^3$ denotes the center of the $i$-th radial basis function, $\sigma > 0$ is the width parameter controlling the spread of each basis function, and $n$ is the number of neurons, which is adaptively determined during training in MATLAB.

Once the RBFNN has been trained, it outputs the predicted correction vector as:
\begin{equation}
    f_\text{RBF}(\hat{\mathbf{X}}) = \Delta \mathbf{X} = [\Delta y, \Delta z, \Delta \theta],
\label{eq:RBFNN3}
\end{equation}
which is then used to compute the corrected state estimates:
\begin{equation}
\left\{
\begin{aligned}
\hat{y}^\text{corr} &= \hat{y} + \Delta y =  \sum_{i=1}^{n} l_i \cos\left( \sum_{j=1}^{i} \hat{\theta}_j \right) + \Delta y, \\
\hat{z}^\text{corr} &= \hat{z} + \Delta z =  \sum_{i=1}^{n} l_i \sin\left( \sum_{j=1}^{i} \hat{\theta}_j \right) + \Delta z, \\
\hat{\theta}^\text{corr} &= \hat{\theta} + \Delta \theta =  \sum_{i=1}^{n} \hat{\theta}_i + \Delta \theta.
\end{aligned}
\right.
\label{eq:RBFNN4}
\end{equation}

It is important to note that the ground-truth data ($y_\text{gt}$, $z_\text{gt}$, $\theta_\text{gt}$) obtained from the Leica Absolute Laser Tracker AT960-LR is utilized only once during the training phase of the RBFNN and is not required during subsequent operation. After training, the proposed system relies solely on the estimated variables ($\hat{y}$, $\hat{z}$, $\hat{\theta}$) to predict the correction terms ($\Delta y$, $\Delta z$, $\Delta \theta$).

\section{Experimental Validation and Results}
\label{sec:experimental}

In this section, the proposed framework is experimentally validated on a physical robotic platform. The description of the experimental setup is first presented, followed by the evaluation results, which demonstrate the effectiveness of the method in practical operation.

\subsection{Experimental Setup}
\label{subsec:experimental}

\begin{figure}[h]
    \centering
    \includegraphics[width=1\linewidth]{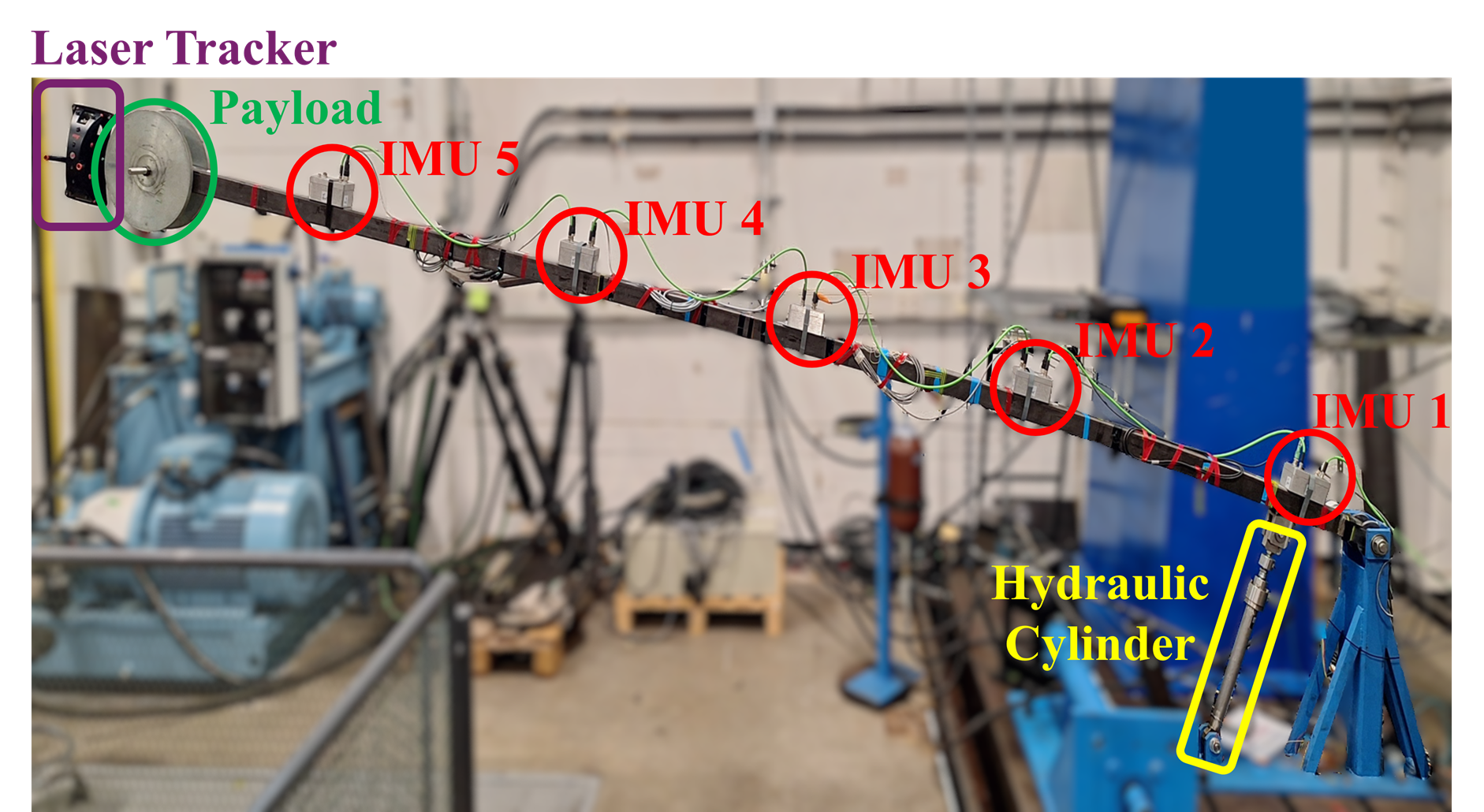}
    \caption{Experimental setup of the flexible manipulator.}
    \label{Experimental Platform}
\end{figure}

The experimental arrangement of the flexible robotic manipulator is illustrated in Fig.~\ref{Experimental Platform}. The system comprises several key subsystems, each fulfilling a distinct function, as described below.  

\subsubsection{Flexible Link}  
The manipulator’s flexible link is fabricated from OPTIM 700 MH Plus, a high-strength structural steel grade. The beam is 4.5~m in length with a square hollow cross-section of 60~mm~$\times$~60~mm and a wall thickness of 3~mm, giving it a mass of approximately 22.5~kg. An additional payload is mounted at the end of the link. The chosen steel offers a yield strength of 700~MPa and an ultimate tensile strength in the range of 750--950~MPa.  

\subsubsection{Hydraulic Actuation System}  
Angular motion of the link is actuated through a hydraulic system incorporating a single-acting hydraulic cylinder ($\oslash$35/25--300~mm) paired with a Bosch Rexroth 4WRPEH electro-hydraulic servo valve, enabling precise control of the link’s position.  

\subsubsection{Distributed IMU Network}  
Kinematic measurements along the link are obtained from a network of five Murata SCHA63T-K03 IMUs positioned at specific locations. This configuration enables continuous monitoring of the link’s motion during vertical movements.  

\subsubsection{Ground-Truth Measurement via Laser Tracker}  
Reference position and orientation data at the end-effector are provided by a Leica Absolute Laser Tracker AT960-LR. This advanced metrology instrument offers sub-millimeter accuracy for real-time tracking of the target in three-dimensional space, ensuring high-quality calibration and validation of the proposed estimation framework.

\subsection{Performance Assessment of RBFNN-Based Kinematic Correction}
\label{subsec:evaluation}

Using the filter parameters optimized in Section~\ref{sec:filter}, a dataset of 84{,}000 samples was collected from experiments covering a wide range of robot motions, with a sampling interval of $0.001~\text{s}$. This dataset was then used to train the RBFNN in MATLAB to predict the residual discrepancy between the actual end-effector position and orientation (obtained from the ground truth) and the estimates provided by the multi-IMU sensor fusion approach as described in Section~\ref{sec:surrogate}. 

To evaluate the effectiveness of the proposed RBFNN-based kinematic correction with 10 neurons, its performance is compared against two baselines: a conventional linear regression (LR) model and the uncorrected estimates (Raw). For a fair comparison, the dataset is divided into training and testing subsets, with a 70\%--30\% split. For a comprehensive comparison of the results, three evaluation metrics are employed: root mean square error (RMSE), mean absolute error (MAE), and maximum error. The results of this evaluation are summarized in Table~\ref{tab:rbfnn_results}.
\begin{table}[h]
\centering
\caption{Performance Comparison of Kinematics Correction Methods}
\label{tab:rbfnn_results}
\begin{tabular}{llccc}
\hline
\textbf{Metric} & \textbf{Method} & \textbf{y [m]} & \textbf{z [m]} & \boldmath{$\theta$ [rad]} \\
\hline
 & Raw    & 0.04406 & 0.03863 & 0.02162 \\
RMSE & LR     & 0.00379 & 0.00629 & 0.00299 \\
 & RBFNN  & \textbf{0.00021} & \textbf{0.00041} & \textbf{0.00024} \\
\hline
  & Raw    & 0.04339 & 0.03817 & 0.02158 \\
MAE  & LR     & 0.00283 & 0.00502 & 0.00223 \\
  & RBFNN  & \textbf{0.00016} & \textbf{0.00033} & \textbf{0.00018} \\
\hline
  & Raw    & 0.05556 & 0.04517 & 0.02393 \\
Max Error  & LR     & 0.01903 & 0.02259 & 0.01747 \\
  & RBFNN  & \textbf{0.00102} & \textbf{0.00156} & \textbf{0.00124} \\
\hline
\end{tabular}
\end{table}

Table~\ref{tab:rbfnn_results} shows that the proposed RBFNN-based kinematic estimation method outperforms both the raw measurements and the linear regression approach. It achieves the lowest RMSE, MAE, and maximum errors for $y$, $z$, and $\theta$, reflecting more precise and reliable position and orientation estimates. Specifically, the RMSE values for RBFNN are $0.00021~\text{[m]}$, $0.00041~\text{[m]}$, and $0.00024~\text{[rad]}$ for $y$, $z$, and $\theta$, respectively. These results indicate that the RBFNN effectively models and compensates for the nonlinear residual errors.

\subsection{Results Before and After RBFNN-Based Kinematic Correction}
\label{subsec:results}

\begin{figure}[h]
    \centering
    \includegraphics[width=1\linewidth]{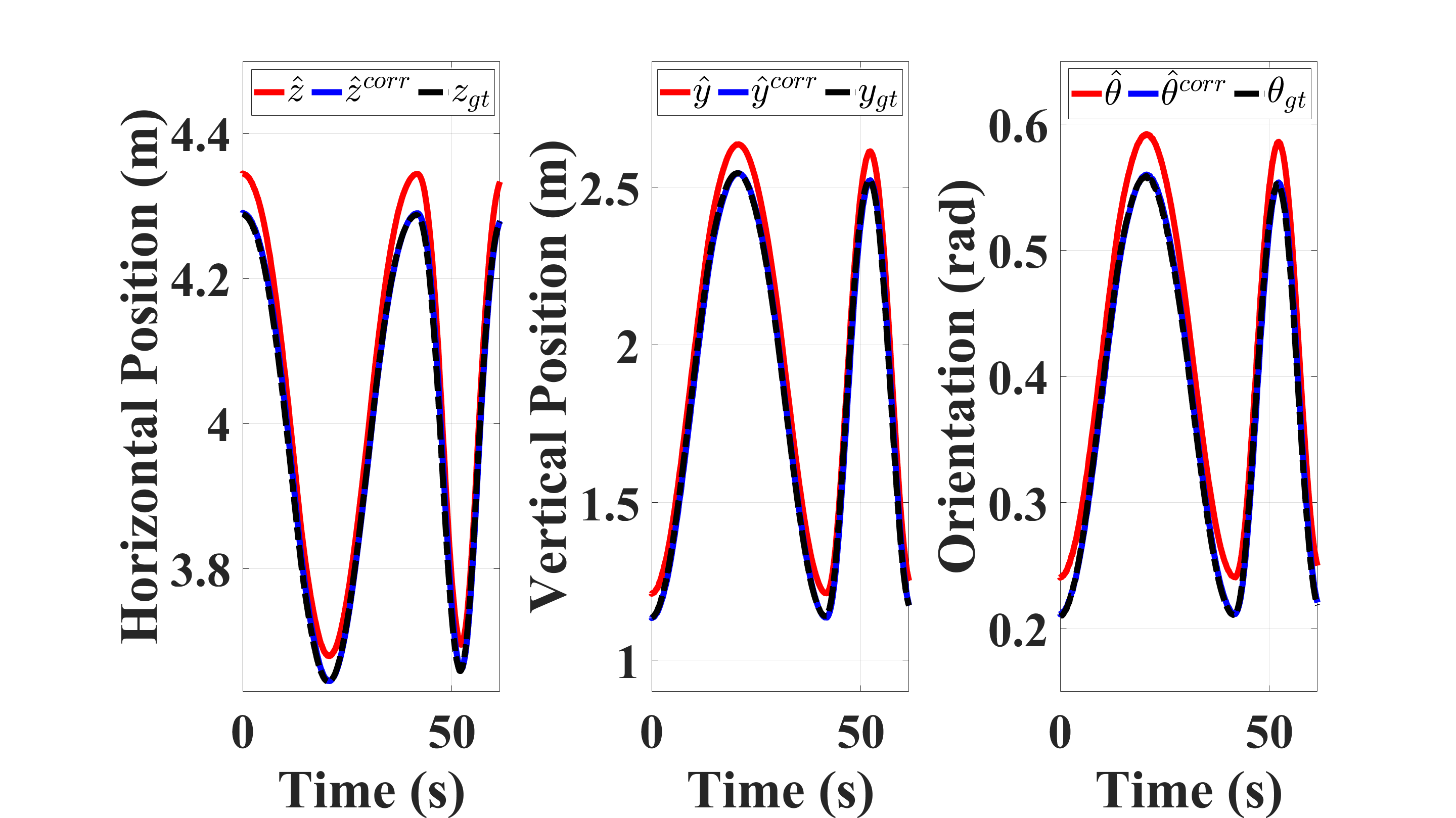}
    \caption{Experimental results with no payload under two operating conditions: $\omega = 0.1$ rad/s (first interval) and $\omega = 0.2$ rad/s (second interval).}
\label{fig:0kg}
\end{figure}

\begin{figure}[h]
    \centering
    \includegraphics[width=1\linewidth]{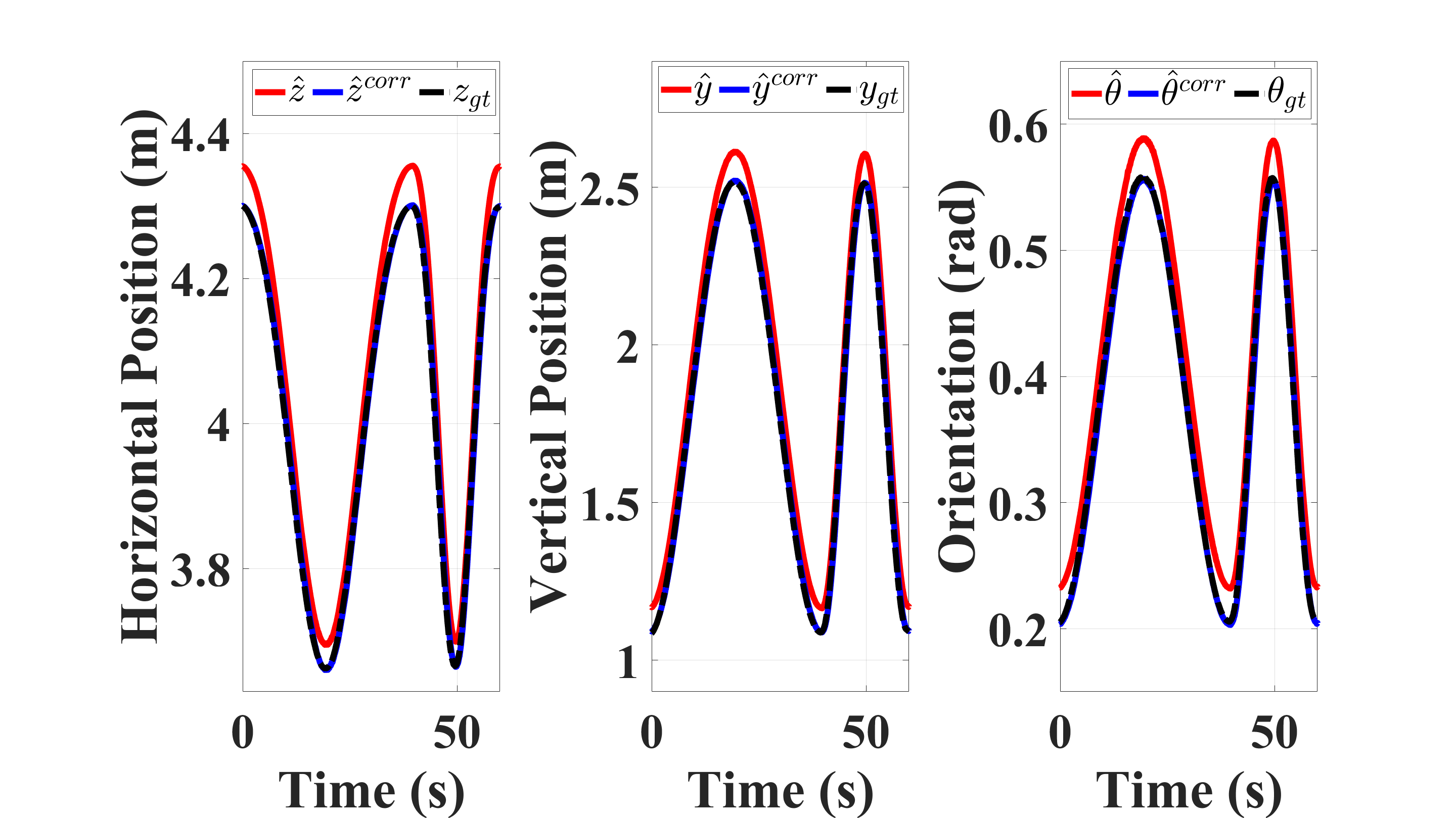}
    \caption{Experimental results with 20 kg payload under two operating conditions: $\omega = 0.1$ rad/s (first interval) and $\omega = 0.2$ rad/s (second interval).}
\label{fig:20kg}
\end{figure}

\begin{figure}[h]
    \centering
    \includegraphics[width=1\linewidth]{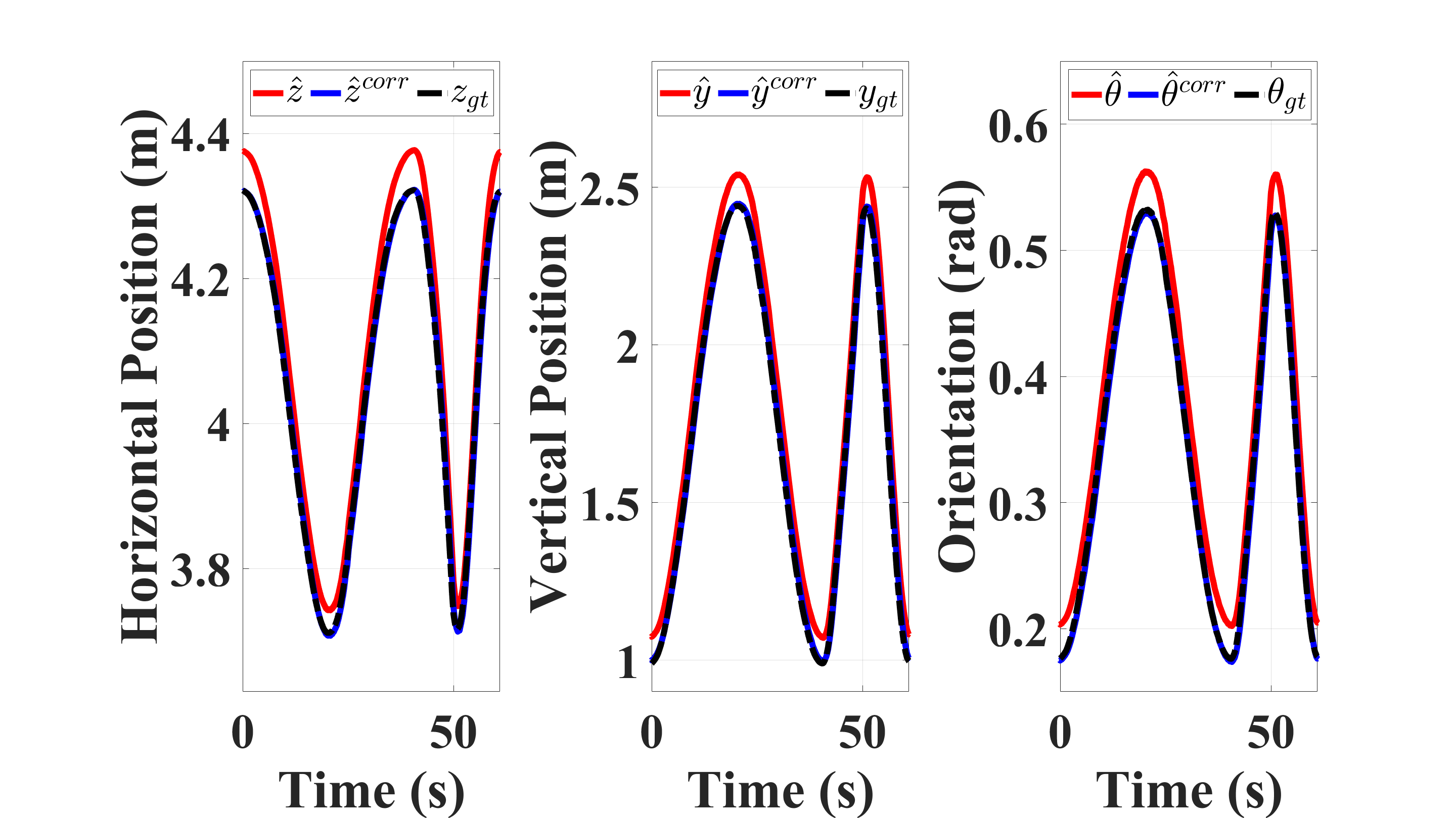}
    \caption{Experimental results with 40 kg payload under two operating conditions: $\omega = 0.1$ rad/s (first interval) and $\omega = 0.2$ rad/s (second interval).}
\label{fig:40kg}
\end{figure}

After training the RBFNN using the ground-truth dataset described in Section~\ref{subsec:evaluation}, its performance is evaluated in real time on the actual robotic system (Fig.~\ref{Experimental Platform}) under a variety of operating conditions. Specifically, the flexible link is tested with different payload configurations (no payload, $20~\mathrm{kg}$, and $40~\mathrm{kg}$) while executing a sinusoidal trajectory at angular velocities of $0.1~\mathrm{rad/s}$, and $0.2~\mathrm{rad/s}$.
The results are shown in Figs.~\ref{fig:0kg}, \ref{fig:20kg}, and \ref{fig:40kg}, where $y$ and $z$ represent the end-effector positions, and $\theta$ represents its orientation, all expressed in the inertial coordinate system depicted in Fig.~\ref{fig:flexible_link}.

It was observed that the proposed RBFNN considerably enhanced the performance of kinematic estimation. Prior to correction, the estimated end-effector positions and orientation ($\hat{y}$, $\hat{z}$, and $\hat{\theta}$) exhibited noticeable deviations from the ground truth data obtained via the laser tracker ($y_\text{gt}$, $z_\text{gt}$, $\theta_\text{gt}$), especially at higher angular velocities and with larger payloads. After applying the RBFNN-based correction, the results ($\hat{y}^\text{corr}$, $\hat{z}^\text{corr}$, and $\hat{\theta}^\text{corr}$) showed substantial improvements, demonstrating superior accuracy and robustness.

\section{Conclusion and Future Work}
\label{sec:conclusions}
This paper introduced an AI-augmented kinematic modeling framework for flexible manipulators, leveraging multi-IMU sensor fusion in conjunction with complementary filtering, PSO-based parameter optimization, and residual error correction via an RBFNN. By integrating IMU-based modeling with data-driven correction, the proposed approach achieved accurate kinematic estimation for vertical motions under gravitational influence, and is applicable to rigid, flexible, and soft robotic systems. The flexible link is represented as a series of rigid segments, with joint angles obtained from low-cost IMUs. Measurement noise, drift, and delay are mitigated through a complementary filter whose parameters are optimally tuned via PSO. The RBFNN further compensated for residual discrepancies between IMU-based estimates and ground-truth measurements, enhancing both positional and orientational accuracy without the need for high-cost or environmentally sensitive sensors. Since only IMU data was required during operation after training, the framework offers a practical solution for industrial applications.

Future work will aim to extend the framework to multi-link flexible manipulators operating in three-dimensional workspaces. Additionally, integration of the proposed IMU-based kinematic modeling with PDE-based dynamic models will be explored to incorporate system physics into the estimation process. Real-time implementation within control loops will also be investigated to enable advanced feedback control strategies for flexible and soft robots.

\bibliographystyle{IEEEtran}
\bibliography{MyReferences.bib}

\end{document}